\documentclass[10pt,twocolumn,letterpaper]{article}

\usepackage{iccv}              % To produce the CAMERA-READY version
\usepackage{multirow}
\usepackage{wrapfig}
\usepackage{appendix}
\usepackage[normalem]{ulem}
\usepackage{graphicx}
\usepackage{amsmath}
\usepackage[accsupp]{axessibility}  % Improves PDF readability for those with disabilities.

\definecolor{iccvblue}{rgb}{0.21,0.49,0.74}
\usepackage[pagebackref,breaklinks,colorlinks,allcolors=iccvblue]{hyperref}

\newcommand{\coolname}{$\texttt{FAPrompt}$\xspace}

\title{Fine-grained Abnormality Prompt Learning for Zero-shot Anomaly Detection}

\author{
  \textbf{Jiawen Zhu}\textsuperscript{1} \quad
  \textbf{Yew-Soon Ong}\textsuperscript{2} \quad
  \textbf{Chunhua Shen}\textsuperscript{3} \quad
  \textbf{Guansong Pang}\textsuperscript{1}\thanks{Corresponding author: Guansong Pang (\texttt{gspang@smu.edu.sg})} \\
  \textsuperscript{1}Singapore Management University, Singapore\\
  \textsuperscript{2}Nanyang Technological University, Singapore\\
  \textsuperscript{3}Zhejiang University, China
}

\begin{document}
\maketitle

\begin{abstract}
Current zero-shot anomaly detection (ZSAD) methods show remarkable success in prompting
large pre-trained vision-language models to detect anomalies in a target dataset without using any dataset-specific training or demonstration. 
However, these methods often focus on crafting/learning prompts that capture only coarse-grained semantics of abnormality, \eg, high-level semantics like `\texttt{damaged}', `\texttt{imperfect}', or `\texttt{defective}' objects. They therefore have limited capability in recognizing diverse abnormality details that deviate from these general abnormal patterns in various ways.
To address this limitation, we propose \textbf{\coolname}, a novel framework designed to 
learn \underline{F}ine-grained \underline{A}bnormality \underline{Prompt}s for accurate ZSAD. 
To this end, a novel \textit{Compound Abnormality Prompt learning} (\textbf{CAP}) module is introduced in \texttt{FAPrompt} to learn a set of complementary, decomposed abnormality prompts, where abnormality prompts are enforced to model diverse abnormal patterns derived from the same normality semantic.
On the other hand, the fine-grained abnormality patterns can be different from one dataset to another. To enhance the cross-dataset generalization,  another novel module, namely \textit{Data-dependent Abnormality Prior learning} (\textbf{DAP}), is introduced in \coolname to learn a sample-wise abnormality prior from abnormal features of each test image to dynamically adapt the abnormality prompts to individual test images.
Comprehensive experiments on 19 real-world datasets, covering both industrial defects and medical anomalies, demonstrate that \coolname substantially outperforms state-of-the-art methods in both image- and pixel-level ZSAD tasks. Code is available at \renewcommand\UrlFont{\color{blue}}\url{https://github.com/mala-lab/FAPrompt}.

\end{abstract}
\vspace{-0.5cm}
\section{Introduction}

Anomaly Detection (AD) is a critical task in computer vision, aiming to identify instances that deviate significantly from the majority of data. It has a wide range of real-world applications, \eg, industrial inspection and medical imaging analysis~\citep{pang2021deep,cao2024survey}. Traditional AD methods
often rely on application-specific, carefully curated datasets to train a specialized detection model, making them inapplicable for application scenarios where such data access is not possible due to data privacy issue, or where the test data significantly differs from the training set due to substantial distribution shifts arising from new deployment environments or other natural variations in datasets.
To address this, Zero-shot Anomaly Detection (ZSAD) has emerged as a promising solution, enabling models to detect anomalies in unseen datasets without dataset-specific training.

\begin{figure}[t]
    \centering
    \includegraphics[width=0.95\linewidth]{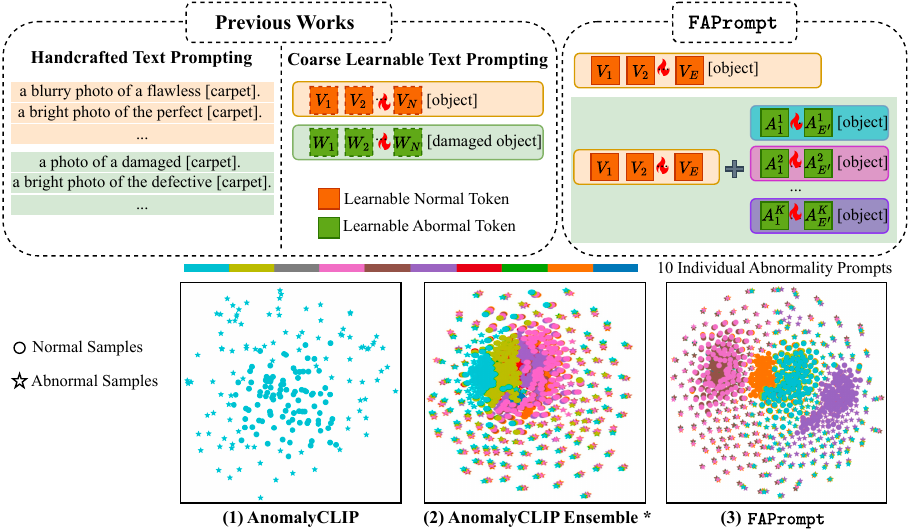}
    \vspace{-0.1cm}
    \caption{\textbf{Top: }\coolname vs. previous methods. \textbf{Bottom}: 2-D t-SNE visualization of the prompt-wise anomaly score maps of three methods on samples from a BTAD dataset~\cite{mishra2021vt}, with the maps generated based on the similarities between text prompt embeddings and image embeddings, as defined in Eq.~\ref{eq:AD-score}.}
    \label{fig:intro}
    \vspace{-0.5cm}
\end{figure}

In recent years, large pre-trained vision-language models (VLMs) such as CLIP~\citep{radford2021learning} have demonstrated impressive zero/few-shot recognition capabilities across a broad range of vision tasks.
To leverage VLMs for ZSAD, existing methods~\citep{chen2024clipadlanguageguidedstageddualpath, jeong2023winclip, deng2023anovl, zhou2023anomalyclip, gu2024filo} use craft/learn text prompts to extract the normal/abnormal textual semantics from VLMs for matching visual anomalies.
Handcrafted text prompting methods such as WinCLIP~\citep{jeong2023winclip} and AnoVL \citep{deng2023anovl} attempt to capture a range of abnormality semantics for better ZSAD by using a wide variety of pre-defined textual description templates as text prompts.
Others~\citep{zhou2023anomalyclip, zhou2022learning, zhou2022conditional, gu2024filo} employ two learnable text prompts to extract more general-purpose features for distinguishing between normal and abnormal classes, such as AnomalyCLIP~\citep{zhou2023anomalyclip} (see Fig.~\ref{fig:intro} \textbf{top}-left). 

However, existing methods often design/learn prompts taking inspiration from image classification, where the abnormality prompts are rudimentary when directly applied to the AD task, leading to a coarse-grained differentiation between normal and abnormal semantics, \eg, high-level semantic of a `\texttt{damaged}', `\texttt{imperfect}', or `\texttt{defective}' carpet. Such prompt designs overlook the intrinsic relationships between normality and abnormality. In practice, abnormalities often manifest as subtle deviations from normal patterns rather than entirely distinct semantic categories, \eg, color stains on normal texture of carpet. As a result, the previous methods are prone to learn excessively simple and homogeneous abnormal patterns, as illustrated in Fig.~\ref{fig:intro} \textbf{bottom} (1), making it difficult to model diverse distributions of abnormal samples, thereby constraining their generalization across different datasets.

To tackle these issues, we propose a novel framework, namely \coolname, designed to learn \underline{F}ine-grained \underline{A}bnormality \underline{Prompt}s for accurate ZSAD. In contrast to previous prompting methods, \coolname focuses on learning the prompts that can model diverse, fine-grained abnormality semantics.
To this end, in \coolname, we introduce a \textit{Compound Abnormality Prompt learning} (\textbf{CAP}) module, which learns a set of complementary, decomposed abnormality prompts on top of a normal prompt, as illustrated in Fig.~\ref{fig:intro} \textbf{top}-right. This compound prompting method ensures that all abnormality prompts are semantically adherent to the normality captured in the normal prompt while introducing learnable deviations to capture fine-grained abnormalities. 
To enhance diversity and minimize redundancy among the abnormality prompts, an orthogonal constraint is further introduced to enforce that each abnormality prompt captures a unique aspect of abnormality while maintaining strong discriminability from the normal class.

On the other hand, most existing methods fail to leverage important prior information hidden in each query/test image, limiting the generalizability of text prompts and their adaptability to unseen datasets. CoCoOp~\cite{zhou2022conditional} attempts to address this issue by incorporating instance-conditional information from image token embeddings. However, its design is ineffective for AD due to two reasons. i) It focuses on class-level semantic difference which can incorporate only coarse-grained semantics into the prompt learning. ii) It can introduce noise into abnormality prompts when the input images are normal samples since they lack abnormality-related information, reducing the discriminability of the learned abnormality prompts (see Table \ref{image-prior}). 

To overcome this challenge, we propose a \textit{Data-dependent Abnormality Prior learning} (\textbf{DAP}) module. It learns to derive abnormality features from the most anomalous patches of each query/test image as a sample-wise abnormality prior to dynamically adapt the abnormality prompts in CAP to individual input images. To avoid using noise (normal features) as the abnormality prior, we impose an abnormality prior learning loss to exclude features from the normal training samples. 

The joint force of CAP and DAP enables \coolname to learn adherent yet diverse fine-grained semantics,
as shown in Fig.~\ref{fig:intro} \textbf{bottom} (3). This effect cannot be achieved by  simple ensemble methods that learn a set of abnormality prompts based on different base models, as they can learn redundant prompts, as shown in Fig.~\ref{fig:intro} \textbf{bottom} (2). Additional results and more t-SNE visualization with ensemble methods can be found in Sec.~\ref{ensemble_methods} and \texttt{Appendix} C.3.

Accordingly, we make the following main contributions.
\begin{itemize}
\item We propose a novel ZSAD framework \coolname. Unlike existing methods that capture coarse-grained semantics of abnormality only, \coolname offers an effective approach for learning diverse fine-grained abnormality semantics.

\item To achieve this, we introduce two novel modules -- CAP and DAP in \coolname. CAP learns a small set of complementary, decomposed fine-grained abnormality prompts via a compounding normal-abnormal token design and an orthogonal constraint, whereas DAP aims to enhance cross-dataset generalization by learning sample-wise abnormality prior from the most abnormal features of abnormal training samples while refraining the prior being affected by normal training images.

\item Comprehensive experiments on 19 diverse real-world industrial and medical image AD datasets show that \coolname significantly outperforms state-of-the-art (SotA) ZSAD models  by at least 3\%-5\% improvement in both image- and pixel-level detection tasks.
\end{itemize}

\vspace{-0.1cm}
\section{Related Work}
\vspace{-0.1cm}
\subsection{Conventional Anomaly Detection}
There have been different types of AD approaches introduced over the years~\cite{pang2021deep,wu2024deep,cao2024survey,zhu2024anomaly}. In particular, one-class classification methods~\citep{tax2004support, yi2020patch, bergman2020classification, chen2022deep, ruff2020deep} aim to compactly describe normal data using support vectors. Reconstruction-based methods~\citep{akcay2019ganomaly, schlegl2019f, zavrtanik2021reconstruction, yan2021learning, zaheer2020old, zavrtanik2021draem, park2020learning, hou2021divide, xiang2023squid, liu2023diversity, yao2023one, yao2023focus} train models to reconstruct normal images, with anomalies identified through higher reconstruction errors. Distance-based methods~\citep{pang2018learning,defard2021padim, cohen2020sub, roth2022towards} detect anomalies by measuring the distance between the test image and normal patterns. Knowledge distillation methods~\citep{deng2022anomaly, bergmann2020uninformed, salehi2021multiresolution, wang2021student, Cao_2023_ICCV, tien2023revisiting, zhang2023destseg} focus on distilling normal patterns from pre-trained models and detecting anomalies by comparing discrepancies between the distilled and original features. However, these methods often rely on application-specific datasets to train the detection model, limiting their applicability in real-world scenarios where data access is restricted due to privacy/resource issues, or where the distribution of test data significant differs from that of the training data.

\vspace{-0.1cm}
\subsection{Zero-Shot Anomaly Detection}
ZSAD has been made possible due to the development of large pre-trained foundation models, such as vision-language models (VLMs). CLIP~\citep{radford2021learning} has been widely used as a VLM to enable ZSAD on visual data \citep{jeong2023winclip,zhou2023anomalyclip,deng2023anovl,chen2023april, cao2024adaclip}. CLIP-AC adapts CLIP for ZSAD by using text prompts designed for the ImageNet dataset as in ~\citep{radford2021learning}. By using manually defined textual prompts specifically designed for industrial AD dataset, WinCLIP~\citep{jeong2023winclip} achieves better ZSAD performance compared to CLIP-AC, but it often does not generalize well to non-defect AD datasets. 

APRIL-GAN~\citep{chen2023april} adapts CLIP to ZSAD through tuning some additional linear layers with annotated auxiliary AD data. AnoVL~\citep{deng2023anovl} introduces domain-aware textual prompts and test time adaptation in CLIP to enhance the ZSAD performance. AnomalyCLIP~\citep{zhou2023anomalyclip}, AdaCLIP~\cite{cao2024adaclip} and FiLO~\cite{gu2024filo} employ learnable textual/visual prompts to extract more general-purpose features for the normal and abnormal classes. All these methods are focused on crafting/learning prompts that capture only coarse-grained semantics of abnormality, failing to detect anomalies that exhibit different patterns from these coarse abnormal patterns. There are a number of other studies leveraging CLIP for AD, but they are designed for empowering few-shot \citep{gu2023anomalyagpt,zhu2024generalistanomalydetectionincontext, li2024promptad} or conventional AD task \citep{joo2023clip,wu2024open,wu2024vadclip,wu2024weakly}.

\begin{figure*}[t]
    \centering
    \includegraphics[width=0.90\linewidth]{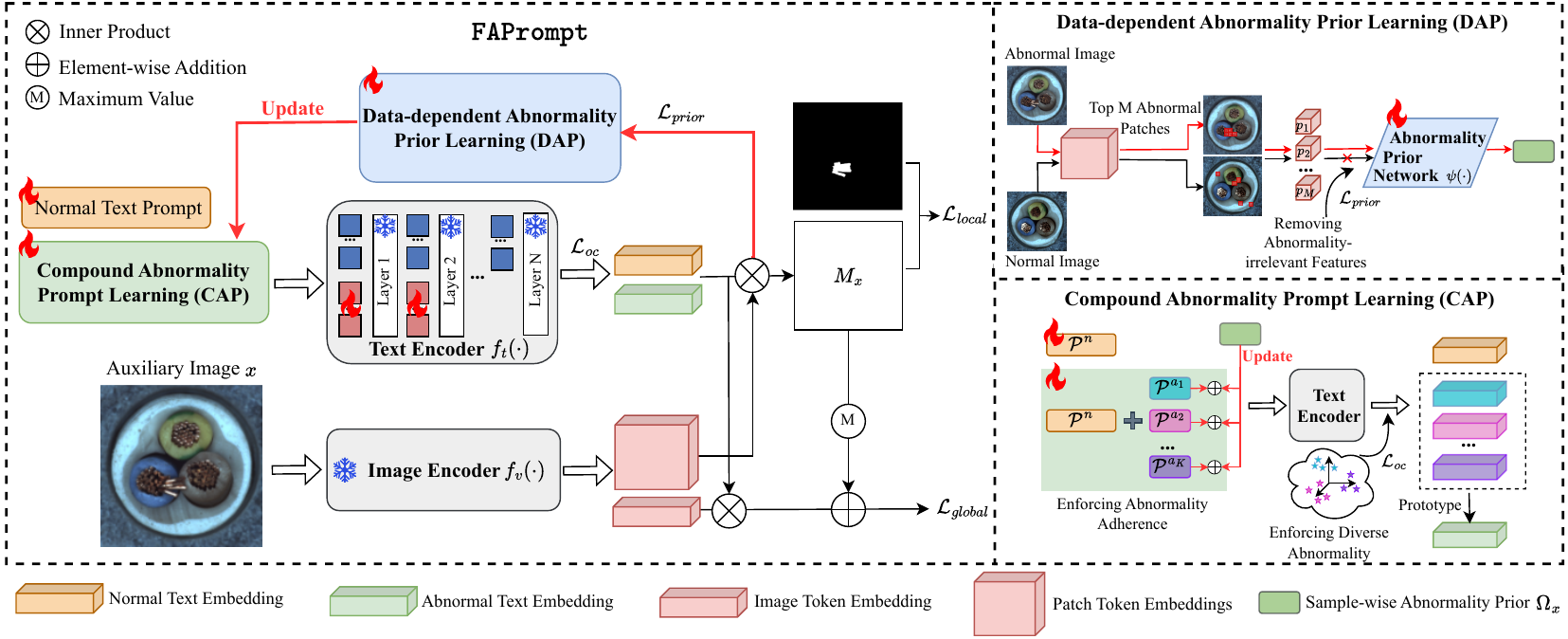}
    \vspace{-0.2cm}
    \caption{Overview of \coolname. It consists of two novel modules: the Compound Abnormality Prompt learning (CAP) (bottom-right) and the Data-dependent Abnormality Prior learning (DAP) (top-right). CAP is devised to learn fine-grained abnormality semantics without relying on detailed human annotations or text descriptions, while DAP is used to adaptively select the most abnormal features from each query/test image as a sample-wise abnormality prior to enhance the cross-dataset generalizability of the abnormality prompts in CAP.
    }
    \label{fig:overall}
    \vspace{-0.4cm}
\end{figure*}

\section{Methodology}
\subsection{Preliminaries}
\textbf{Problem Statement.}
Let $\mathcal{D}_{train} = \{X_{train}, Y_{train}\}$ denote an auxiliary training dataset consisting of both normal and anomalous samples, where $X_{train} = \{x_i\}_{i=1}^{N}$ is a set of $N$ images and $Y_{train} = \{y_i, \mathbf{G}_i\}_{i=1}^{N}$ contains the corresponding ground truth labels and pixel-level anomaly masks. Each image $x_i$ is labeled by $y_i$, where $y_i = 0$ indicates a normal image and $y_i = 1$ signifies an anomalous one. The anomaly mask $\mathbf{G}_i$ provides pixel-level annotation of $x_i$.
% , where each pixel have value 0 if normal and 1 if abnormal. 
During the testing phase, we are presented with a collection of target datasets, $\mathcal{T}=\{\mathcal{D}_{test}^1,\mathcal{D}_{test}^2,\cdots,\mathcal{D}_{test}^t\}$, where each $\mathcal{D}_{test}^j = \{X_{test}^j, Y_{test}^j\}$ is a test set from a target application dataset that have different normal and abnormal samples from those in the training data $\mathcal{D}_{train}$. The goal of ZSAD is to develop models on the auxiliary dataset $\mathcal{D}_{train}$, with the ability to generalize to detect anomalies in different test sets in $\mathcal{T}$. Particularly, given an input RGB image $x \in \mathbb{R}^{h \times w \times 3}$ from $\mathcal{D}_{train}$, with $h$ and $w$ respectively representing the height and width of $x$, a ZSAD model is required to output both an image-level anomaly score $s_x \in \mathbb{R}$ and a pixel-level anomaly map $\mathbf{\mathcal{M}}_x \in \mathbb{R}^{h \times w}$. The image-level anomaly score $s_x$ provides a global assessment of whether the image is anomalous, while the pixel-level anomaly map $\mathbf{\mathcal{M}}_x$ indicates the likelihood of each pixel being anomalous. Both $s_x$ and the values in $\mathbf{\mathcal{M}}_x$ lie in [0, 1], where a larger value indicates a higher probability of being abnormal.

\noindent\textbf{VLM Backbone.} To enable accurate ZSAD, large pre-trained VLMs are typically required. Following existing approaches~\citep{deng2023anovl, jeong2023winclip, zhou2023anomalyclip, cao2024adaclip}, we select CLIP~\citep{radford2021learning} in our study, which comprises a visual encoder $f_v(\cdot)$ and a text encoder $f_t(\cdot)$, where the visual and text features are well-aligned through pre-training on web-scaled text-image pairs. 

\vspace{-0.1cm}
\subsection{Overview of \coolname}
In this work, we propose a ZSAD framework \coolname to learn fine-grained abnormality semantics without any reliance on detailed human annotations or text descriptions. Fig~\ref{fig:overall} illustrates the overall framework of \coolname that consists of two novel modules, including Compound Abnormality Prompt learning (\textbf{CAP}) module and Data-dependent Abnormality Prior learning (\textbf{DAP}) module.

To be more specific, the proposed CAP module is devised to specify the design of fine-grained abnormality prompts. The key characteristic of CAP is to obtain the abnormality prompts via a compound prompting method, where we have one normal prompt and multiple abnormality prompts are added on top of it. These normal and abnormal text prompts are then processed by the CLIP’s text encoder $f_t(\cdot)$ to generate the corresponding normal and abnormal text embeddings, respectively. For a given image $x$, \coolname extracts both an image token embedding $f_v(x)$ and a set of patch token embeddings $\mathbf{F}_v \in \mathbb{R}^{l \times d} $, with $l$ and $d$ respectively representing the length and dimensionality of $\mathbf{F}_v$. The prompts are then learned using $\mathcal{D}_{train}$ based on the similarity between the visual and text embeddings, where the fine-grained abnormality prompts are aggregated into an abnormality prompt prototype before its use in similarity calculation. 

Further, the DAP module is introduced to improve cross-dataset generalization capability of the fine-grained abnormality prompts. DAP derives the most relevant abnormality features from a given query/test image $x$, serving as a sample-wise abnormality prior to dynamically adapt the abnormality prompts in CAP to individual samples of a given target dataset. During training, the original parameters of CLIP remain frozen, and only the attached learnable tokens in the text encoder layers, along with the normal and fine-grained abnormality prompts, are optimized. Below we present these modules in detail.

\vspace{-0.1cm}
\subsection{Compound Abnormality Prompt Learning}
\noindent\textbf{Learning Fine-grained Abnormalities via Compound Normal and Abnormal Tokens.}
Previous approaches that rely on coarse-grained learnable text prompts fail to capture the fine-grained abnormality semantics for detecting diverse anomalies across different datasets. To address this, we propose the novel CAP module. The core insight is that abnormal samples typically exhibit different magnitude of deviation from their normal counterparts while still being adherent to the normal class. CAP models this by learning a set of complementary, decomposed abnormality prompts built on a shared normal prompt. 
To be specific, a set of learnable normal tokens and the fixed token `object' are concatenated to define the normal text prompt $\mathcal{P}^n$. For the abnormality prompt, CAP aims to learn a small set of prompts of complementary semantics, denoted as $\mathcal{P}^a = \{\mathcal{P}^{a_1}, \mathcal{P}^{a_2}, ... \mathcal{P}^{a_K}\}$, where each $\mathcal{P}^{a_i}$ is formed by a compound of the same tokens in the normal prompt $\mathcal{P}^n$ and a few learnable abnormal tokens. 
Formally, the normal and abnormality prompts can be defined as:

\vspace{-0.3cm}
\begin{equation}
\begin{aligned}
    &\mathcal{P}^n = [V_1] [V_2] ... [V_{E}] [object], \\
    &\mathcal{P}^a = \{\mathcal{P}^{a_1}, \mathcal{P}^{a_2}, ... \mathcal{P}^{a_K}\}, \\
    with\ \mathcal{P}^{a_i} &= [V_1] [V_2] ... [V_{E}] [A_1^i] [A_2^i] ...[A_{E^\prime}^i] [object],
    \label{eq:prompt}
\end{aligned}
\end{equation}
where $\{V_1, V_2, ... V_{E}\}$ and $\{A_1^i, A_2^i, ... A_{E^\prime}^i\}_{i=1}^K$ are learnable normal and abnormal tokens, respectively. This compound prompting strategy enables the learning of different abnormality semantics easily while maintaining abnormality prompts being adherent to the normality in the normal prompt, supporting the learning of semantically-meaningful abnormality prompts.

\noindent\textbf{Learning Complementary Abnormality Prompts.}
To capture complementary fine-grained abnormalities and reduce redundant information captured by the abnormality prompts, it is essential to maximize the diversity among the fine-grained abnormalities. A straightforward approach would be to train distinct abnormality prompts on separate, annotated subsets with samples from different anomalous types. However, this would require extensive human annotations. To address this issue, we propose to add an orthogonal constraint loss $\mathcal{L}_{oc}$ into the abnormality prompts in CAP as a alternative method to encourage this diversity. 
Formally, the objective for this can be formulated as:

\vspace{-0.2cm}
\begin{equation}
    \mathcal{L}_{oc} = \sum_{i, j \in K; i \neq j}abs(\frac{f_t(\mathcal{P}^{a_i}) \cdot f_t(\mathcal{P}^{a_j})}{||f_t(\mathcal{P}^{a_i})|| \times ||f_t(\mathcal{P}^{a_j})||}),
    \label{eq:dis}
\end{equation}
where the text encoder $f_t(\cdot)$ is used to extract the embeddings of the abnormality prompts, $[\cdot]$ denotes inner product, $abs(\cdot)$ returns the absolute value, and $||\cdot||$ indicates the norm of vectors. To provide more representative embedding for the fine-grained abnormalities, we compute the prototype of the multiple abnormality prompt embeddings as the final fine-grained abnormality embedding via $\mathbf{F}_a = \frac{1}{|\mathcal{P}^a|}\sum_{\mathcal{P}^{a_i} \in \mathcal{P}^a}f_t(\mathcal{P}^{a_i})$. The normal text prompt embedding is $\mathbf{F}_n = f_t(\mathcal{P}^n)$. 

\vspace{-0.1cm}
\subsection{Data-dependent Abnormality Prior Learning}
One issue in ZSAD is that the fine-grained abnormality patterns can be very different from the auxiliary dataset to test datasets. In addition to the learning of a set of complementary fine-grained abnormality prompts, it is important to ensure that the learned fine-grained abnormality patterns are generalized to various test datasets. Inspired by the instance-conditional information design in CoCoOp~\citep{zhou2022conditional}, we introduce the DAP module to enhance the cross-dataset generalizability of the abnormal tokens in CAP by adaptively selecting the embeddings of the most abnormal regions to serve as a sample-wise abnormality prior for each image input. Particularly, given a query/test image $x$, DAP selects the most abnormal image patches as the abnormality prior to be fed into CAP for assisting the abnormality prompt learning. It achieves this by picking the top $M$ patches whose token embeddings are most similar to the abnormality prompt prototype $\mathbf{F}_a$:

\vspace{-0.2cm}
\begin{equation}
    \mathbf{S}_x^a(i, j) = \frac{\exp(\mathbf{F}_v(i, j)\mathbf{F}_a^{\intercal})}{\exp(\mathbf{F}_v(i, j)\mathbf{F}_n^{\intercal}) + \exp(\mathbf{F}_v(i, j)\mathbf{F}_a^{\intercal})},
    \label{eq:AD-score}
\end{equation}
where $[\cdot]^\intercal$ denotes a transpose operation, $\mathbf{F}_v(i, j)$ is the token embedding of the patch centered at $(i,j)$ and $\mathbf{S}_x^a(i, j)$ is a patch-level anomaly score. The corresponding normal scores can be calculated via $\mathbf{S}_x^n(i, j)$ using the similarity to $\mathbf{F}_n$ in the numerator in Eq.~\ref{eq:AD-score}.

Let $\textbf{p}_x = \{p_1, p_2, ...p_M\}$ be the top $M$ patch embeddings of $x$, \coolname then adds additional learnable layers $\psi(\cdot)$, namely \textit{abnormality prior network}, to learn the sample-wise abnormality prior based on $\textbf{p}_x$. This prior $\Omega_x = \psi(\textbf{p}_x)$ is then incorporated as data-dependent abnormal features into the learnable abnormal tokens of the abnormality prompts in CAP to dynamically adapt the fine-grained abnormalities to a given target dataset, with each individual abnormality prompt refined as follows:

\vspace{-0.2cm}
\begin{equation}
    \mathcal{\hat{P}}^{a_i} = [V_1] ... [V_{E}] [A_1^i \oplus\Omega_x]...[A_{E^\prime}^i\oplus\Omega_x] [object],
    \label{eq:new_prompt}
\end{equation}
where $\Omega_x$ is a vector-based prior of the same dimensionality as the abnormal tokens and $\oplus$ denotes element-wise addition. Thus, the abnormality prompt set is updated as $\mathcal{\hat{P}}^a = \{\mathcal{\hat{P}}^{a_1}, \mathcal{\hat{P}}^{a_2}, ... \mathcal{\hat{P}}^{a_K}\}$, and the abnormality prompt prototype can be accordingly refined as 
$\mathbf{\hat{F}}_a = \frac{1}{|\mathcal{\hat{P}}^a|}\sum_{\mathcal{\hat{P}}^{a_i} \in \mathcal{\hat{P}}^a}f_t(\mathcal{\hat{P}}^{a_i})$. 

The goal of DAP is to introduce sample-wise \textit{abnormality} information. However, there is no abnormality information from the top $M$ patches of normal images, and thus, simply applying the prior $\Omega_x$ to normal images would
introduce noise into the learnable abnormal tokens, damaging the learning of fine-grained abnormalities. To address this issue, we propose an abnormality prior learning loss $\mathcal{L}_{prior}$ to enforce that $\Omega_x$ is the features mapped from the most abnormal $M$ patches if $x$ is an abnormal image, while it is minimized to be a null vector if it is a normal image. Formally, $\mathcal{L}_{prior}$ can be defined as follows:

\begin{equation}
    \mathcal{L}_{prior} = \sum_{y_x = 0}\sum_{\omega \in \Omega_x}\omega_x^2,
    \label{eq:filter}
\end{equation}
where $\omega$ is an entry of $\Omega_x$. 

\vspace{-0.1cm}
\subsection{Training and Inference}

\noindent\textbf{Training.}
During training, \coolname first generates an abnormality-oriented segmentation map $\mathbf{\mathcal{\hat{M}}}^a \in \mathbb{R}^{h \times w}$ using ${\hat{S}_x^a}$ whose entries are calculated via Eq.~\ref{eq:AD-score} with $\mathbf{F}_a$ replaced by the prior-enabled $\mathbf{\hat{F}}_a$:

\vspace{-0.2cm}
\begin{equation}
 \mathbf{\mathcal{\hat{M}}}^a= \Phi({\hat{S}_x^a}),
    \label{eq:maps}
\end{equation}
where $\Phi(\cdot)$ is a reshape and interpolation function that transforms the patch-level anomaly scores into a two-dimensional segmentation map. In the same way, we can generate the segmentation map $\mathbf{\mathcal{\hat{M}}}^n = \Phi({\hat{S}_x^n})$ based on the prior-enabled normal score ${\hat{S}_x^n}$. Let $\mathbf{G}_x$ represent the ground-truth mask of the query image $x$, following previous works\citep{zhou2023anomalyclip, chen2023april}, the learning objective in \coolname for optimizing pixel-level AD can then be defined as:

\vspace{-0.2cm}
\begin{equation}
\begin{split}
    \mathcal{L}_{local} & = \frac{1}{N}\sum_{x\in X_{train}}\mathcal{L}_{Focal}([\mathbf{\mathcal{\hat{M}}}^n, \mathbf{\mathcal{\hat{M}}}^a], \mathbf{G}_x)\\
    & + \mathcal{L}_{Dice}(\mathbf{\mathcal{\hat{M}}}^a, \mathbf{G}_x) + \mathcal{L}_{Dice}(\mathbf{\mathcal{\hat{M}}}^n, \textbf{I} - \mathbf{G}_x),
    \label{eq:local}
\end{split}
\end{equation}
where $\textbf{I}$ is a full-one matrix, $\mathcal{L}_{Focal}(\cdot)$ and $\mathcal{L}_{Dice}(\cdot)$ denote a focal loss~\citep{lin2018focallossdenseobject} and a dice loss~\citep{li2020dicelossdataimbalancednlp}, respectively. To ensure the accuracy of locating the top abnormal features in DAP, we apply the same learning objective to optimize the segmentation maps $\mathbf{\mathcal{M}^n} \in \mathbb{R}^{h \times w}$ and $\mathbf{\mathcal{M}^a} \in \mathbb{R}^{h \times w}$, which are derived from the normality-oriented scores ${S_x^n}$ and abnormality-oriented scores ${S_x^a}$, respectively. 

For image-level supervision, \coolname first computes the probability of the query image $x$ being classified as abnormal based on its cosine similarity to the two prompt embeddings $\mathbf{\hat{F}}_a$ and $\mathbf{F}_n$:

\vspace{-0.2cm}
\begin{equation}
    s_a(x) = \frac{\exp(f_v(x) \mathbf{\hat{F}}_a^{\intercal})}{\exp(f_v(x)\mathbf{F}_n^{\intercal}) + \exp(f_v(x)\mathbf{\hat{F}}_a^{\intercal})}.
    \label{eq:score}
\end{equation}

The final image-level anomaly score is then defined as the average of this image-level score and the maximum pixel-level anomaly score derived from the anomaly score maps:

\vspace{-0.2cm}
\begin{equation}
    s(x) = \frac{1}{2}(s_a(x) + s_a^\prime(x)),
    \label{eq:finalscore}
\end{equation}
where $s_a^\prime(x)=\frac{1}{2}\left(\max({S_x^a}) + \max({\hat{S}}_x^a)\right)$ represents the average of the maximum anomaly scores from ${S_x^a}$ and ${\hat{S}_x^a}$. Following previous methods~\citep{zhu2024generalistanomalydetectionincontext, chen2023april, zhou2023anomalyclip, jeong2023winclip}, $s_a^\prime(x)$ is treated as a complementary anomaly score to $s_a(x)$ and incorporated into Eq.~\ref{eq:finalscore}, as $s_a^\prime(x)$ are helpful for detecting local abnormal regions.
The image-level anomaly score $s(x)$ is then optimized by minimizing the following loss on $X_{train}$:

\vspace{-0.2cm}
\begin{equation}
    \mathcal{L}_{global} = \frac{1}{N}\sum_{x\in X_{train}}\mathcal{L}_{b}(s(x), y_x),
    \label{eq:global}
\end{equation}
where $\mathcal{L}_{b}$ is specified by a focal loss function due to the class imbalance in $X_{train}$.
Overall, \coolname is optimized by minimizing the following combined loss, which integrates both local and global objectives, along with the two constraints from the CAP and DAP modules:

\begin{table*}[t]
\centering
\resizebox{0.95\textwidth}{!}{
\begin{tabular}{c|c|ccccc|cccccc}
\toprule
\multirow{2}{*}{\textbf{Data Type}} & \multirow{2}{*}{\textbf{Dataset}} & \multicolumn{5}{c|}{\textbf{Handcrafted Text Prompts}}                                    & \multicolumn{6}{c}{\textbf{Learnable Text Prompts}}                          \\
                                    &                                       & \textbf{CLIP} & \textbf{CLIP-AC} & \textbf{WinCLIP} & \textbf{APRIL-GAN} & \textbf{AnoVL} & \textbf{CoOp} & \textbf{CoCoOp}  &  \textbf{BLIP}         & \textbf{AnomalyCLIP} & \textbf{FiLO} & \textbf{\coolname} \\ \midrule
\multirow{9}{*}{\textbf{Industrial}}           & \textbf{MVTecAD}                      & (74.1, 87.6)  & (71.5, 86.4)     & (91.8, {\color[HTML]{0000FF}\textbf{96.5}})     & (86.2, 93.5)       & ({\color[HTML]{FF0000}\textbf{92.5}}, {\color[HTML]{FF0000}\textbf{96.7}})   & (88.8, 94.8)  & (71.8, 84.9) & (72.3, 86.7) & (91.5, 96.2)  & (91.2, 96.0)         & ({\color[HTML]{0000FF}\textbf{91.9}}, 95.7)  \\ 
   & \textbf{VisA}                         & (66.4, 71.4)  & (65.0, 70.2)     & (78.8, 81.4)     & (78.0, 81.4)       & (79.2, 81.7)   & (62.8, 68.1)  & (78.1, 82.3)    & (57.8, 64.6)     & (82.1, 84.6)  & ({\color[HTML]{0000FF}\textbf{83.9}}, {\color[HTML]{FF0000}\textbf{87.3}})         & ({\color[HTML]{FF0000}\textbf{84.6}}, {\color[HTML]{0000FF}\textbf{86.8}})  \\
                                    & \textbf{BTAD}                         & (34.5, 52.5)  & (51.0, 62.1)     & (68.2, 70.9)     & (73.6, 68.6)       & (80.3, 73.1)   & (66.8, 77.4)  & (48.4, 53.9)      & (85.3, 91.2)   & ({\color[HTML]{0000FF}\textbf{88.3}}, 87.3)  & (85.4, {\color[HTML]{0000FF}\textbf{91.5}})         & ({\color[HTML]{FF0000}\textbf{92.2}}, {\color[HTML]{FF0000}\textbf{92.5}})  \\ 
%                                     \hline
% \multirow{4}{*}{\textbf{Textual}}  
& \textbf{AITEX}                        & (71.0, 45.7)  & (71.5, 46.7)     & (73.0, 54.7)     & (57.6, 41.3)       & (72.5, 55.4)   & (66.2, 39.0)  & (48.6, 37.8)  & (71.3, 52.2)       & (62.2, 40.4)  & ({\color[HTML]{FF0000}\textbf{78.2}}, {\color[HTML]{FF0000}\textbf{59.9}})         & ({\color[HTML]{0000FF}\textbf{74.1}}, {\color[HTML]{0000FF}\textbf{55.5}})  \\
                                    & \textbf{DAGM}                         & (79.6, 59.0)  & (82.5, 63.7)     & (91.8, 79.5)     & (94.4, 83.8)       & (89.7, 76.3)   & (87.5, 74.6)  & (96.3, 85.5) & (84.3, 66.6)        & ({\color[HTML]{0000FF}\textbf{97.5}}, {\color[HTML]{0000FF}\textbf{92.3}})   &  (96.6, 90.4)       & ({\color[HTML]{FF0000}\textbf{98.8}}, {\color[HTML]{FF0000}\textbf{95.3}})  \\
                                    & \textbf{DTD-Synthetic}                & (71.6, 85.7)  & (66.8, 83.2)     & (93.2, 92.6)     & (86.4, 95.0)       & ({\color[HTML]{0000FF}\textbf{94.9}}, 97.3)   & (83.1, 91.9)  & (84.1, 92.9)   & (91.5, 96.3)      & (93.5, 97.0)  & (94.7, {\color[HTML]{0000FF}\textbf{98.0}})         & ({\color[HTML]{FF0000}\textbf{96.2}}, {\color[HTML]{FF0000}\textbf{98.1}})  \\
                                    & \textbf{ELPV}                         & (59.2, 71.7)  & (69.4, 80.2)     & (74.0, 86.0)     & (65.5, 79.3)       & (70.6, 83.0)   & (73.0, 86.5)   &     (78.4, 89.2) & (75.2, 87.3)   & (81.5, {\color[HTML]{0000FF}\textbf{91.3}})  & ({\color[HTML]{0000FF}\textbf{82.2}}, 91.2)         & ({\color[HTML]{FF0000}\textbf{83.7}}, {\color[HTML]{FF0000}\textbf{92.1}})  
                                    
                                    \\
                                    & \textbf{SDD}                          & (95.5, 87.9)  & (94.7, 77.9)     & (94.0, 87.2)     & (97.5, 93.4)       & (95.3, 91.3)   & (96.8, 90.0)  & (89.9, 50.4) & (85.8, 56.9)      & ({\color[HTML]{0000FF}\textbf{98.1}}, 93.4)  & (97.9, {\color[HTML]{0000FF}\textbf{94.8}})         & ({\color[HTML]{FF0000}\textbf{98.4}}, {\color[HTML]{FF0000}\textbf{95.6}})

                                    \\
                                    & \textbf{MPDD}                         & (54.3, 65.4)  & (56.2, 66.0)     & (63.6, 69.9)     & (73.0, 80.2)       & (68.9, 71.9)   & (55.1, 64.2)  & (61.0, 69.1) & (59.5, 67.7)        & ({\color[HTML]{0000FF}\textbf{77.0}}, {\color[HTML]{0000FF}\textbf{82.0}})   & (74.4, 76.9)         & ({\color[HTML]{FF0000}\textbf{80.1}}, {\color[HTML]{FF0000}\textbf{83.9}})
                                    
                                    \\ \midrule
\multirow{4}{*}{\textbf{Medical}}   & \textbf{BrainMRI}                     & (73.9, 81.7)  & (80.6, 86.4)     & (86.6, 91.5)     & (89.3, 90.9)       & (88.7, 91.3)   & (61.3, 44.9)  & (78.2, 86.7)    & (75.7, 83.0)     & (90.3, 92.2)  & ({\color[HTML]{0000FF}\textbf{94.5}}, {\color[HTML]{0000FF}\textbf{94.9}})         & ({\color[HTML]{FF0000}\textbf{95.8}}, {\color[HTML]{FF0000}\textbf{96.2}})  \\
                                    & \textbf{HeadCT}                       & (56.5, 58.4)  & (60.0, 60.7)     & (81.8, 80.2)     & (89.1, 89.4)       & (81.6, 84.2)   & (78.4, 78.8)  & (80.3, 73.4)  & (68.1, 62.8)       & (93.4, 91.6)   & ({\color[HTML]{0000FF}\textbf{93.9}}, {\color[HTML]{0000FF}\textbf{92.1}})         & ({\color[HTML]{FF0000}\textbf{94.0}}, {\color[HTML]{FF0000}\textbf{92.4}})  \\
                                    & \textbf{LAG}                          & (58.7, 76.5)  & (58.2, 76.9)     & (59.2, 74.8)     & (73.6, 84.8)       & (65.1, 78.0)   & (69.6, 82.9)  & (72.6, 84.7) & (48.3, 66.8)        & ({\color[HTML]{0000FF}\textbf{74.3}}, {\color[HTML]{0000FF}\textbf{84.9}})    & (71.3, 83.8)         & ({\color[HTML]{FF0000}\textbf{76.6}}, {\color[HTML]{FF0000}\textbf{86.1}})  \\
                                    & \textbf{Br35H}                        & (78.4, 78.8)  & (82.7, 81.3)     & (80.5, 82.2)     & (93.1, 92.9)       & (88.4, 88.9)   & (86.0, 87.5)  & (85.7, 89.1) & (76.4, 74.4)        & (94.6, 94.7)    & ({\color[HTML]{FF0000}\textbf{97.7}}, {\color[HTML]{0000FF}\textbf{96.8}})         & ({\color[HTML]{0000FF}\textbf{97.6}}, {\color[HTML]{FF0000}\textbf{97.1}})  
                                    
                                    \\ \bottomrule
\end{tabular}
}
\caption{Image-level ZSAD results (AUROC, AP) on 13 AD datasets. The best and second-best results are respectively highlighted in {\color[HTML]{FF0000}\textbf{red}} and {\color[HTML]{0000FF}\textbf{blue}}. The results for MVTecAD, VisA, DAGM, DTD-Synthetic, BTAD, and MPDD are averaged performance across their multiple data subsets (see \texttt{Appendix} D.1 for breakdown results).
}
\label{classification}
\vspace{-0.4cm}
\end{table*}

\noindent\textbf{Inference.}
During inference, given a test image $x^\prime$, it is fed through the \coolname framework to generate the segmentation maps $\mathbf{\mathcal{M}}^n$, $\mathbf{\mathcal{M}}^a$, $\mathbf{\mathcal{\hat{M}}}^n$, and $\mathbf{\mathcal{\hat{M}}}^a$. Then the pixel-level anomaly map $\mathbf{\mathcal{M}}_{x^\prime}$ is calculated by averaging over these segmentation maps as follows:

\vspace{-0.2cm}
\begin{equation}
    \mathbf{\mathcal{M}}_{x^\prime} = \frac{1}{4}(\mathbf{\mathcal{M}}^a \oplus 1 \ominus \mathbf{\mathcal{M}}^n \oplus \mathbf{\mathcal{\hat{M}}}^a \oplus 1 \ominus \mathbf{\mathcal{\hat{M}}}^n),
    \label{eq:map}
\end{equation}
where $\ominus $ is element-wise subtraction. The image-level anomaly score $s_{x^\prime}$ is computed using Eq.~\ref{eq:finalscore}.

\vspace{-0.1cm}
\section{Experiments}

\noindent\textbf{Datasets.} 
To verify the effectiveness of \coolname, we conduct extensive experiments across 19 publicly available datasets, including nine popular industrial defect inspection datasets on varying products/objects (MVTecAD~\citep{bergmann2019mvtec}, VisA~\citep{zou2022spot}, DAGM~\citep{wieler2007weakly}, DTD-Synthetic~\citep{aota2023zero}, AITEX~\citep{silvestre2019public}, SDD~\citep{tabernik2020segmentation}, BTAD~\citep{mishra2021vt}, MPDD~\citep{jezek2021deep}, and ELPV\citep{deitsch2019automatic}) and ten medical anomaly detection datasets on different organs like brain, fundus, colon, skin and thyroid (BrainMRI~\citep{salehi2021multiresolution}, HeadCT~\citep{salehi2021multiresolution}, LAG~\citep{li2019attention}, Br35H~\citep{hamada2020br35h}, CVC-ColonDB~\citep{tajbakhsh2015automated}, CVC-ClinicDB~\citep{bernal2015wm}, Kvasir~\citep{jha2020kvasir}, Endo~\citep{hicks2021endotect}, ISIC~\citep{gutman2016skin}, TN3K~\citep{gong2021multi}).
To assess the ZSAD performance, the models are trained on the MVTecAD dataset by default 
and evaluated on the test sets of other datasets without any further training or fine-tuning. We obtain the ZSAD results on MVTecAD by changing the training data to the VisA dataset
(see \texttt{Appendix} A for details about the datasets).

\noindent\textbf{Competing Methods and Evaluation Metrics.}
We compare our \coolname with several state-of-the-art (SotA) methods, including five handcrafted text prompt-based methods (raw CLIP~\citep{radford2021learning}, CLIP-AC, WinCLIP~\citep{jeong2023winclip}, APRIL-GAN~\citep{chen2023april}, and AnoVL~\citep{deng2023anovl}) and five learnable text prompt-based methods (CoOp~\citep{zhou2022learning}, CoCoOp~\citep{zhou2022conditional}, AnomalyCLIP~\citep{zhou2023anomalyclip}), FiLO~\citep{gu2024filo}, and BLIP~\cite{li2022blip}). As for evaluation metrics, we follow previous works \citep{jeong2023winclip,zhou2023anomalyclip} and use two popular metrics: Area Under the Receiver Operating Characteristic (AUROC) and Average Precision (AP) to assess the image-level AD performance; for pixel-level AD performance, we employ AUROC and Area Under Per Region Overlap (PRO) to provide a more detailed analysis. 
We also provide model complexity analysis and detailed qualitative results in \texttt{Appendix} C.1 and C.5, respectively.

\noindent\textbf{Implementation Details.}
The implementation details for \coolname and the competing methods are provided in \texttt{Appendix} B.

\vspace{-0.3cm}
\begin{table*}[t]
\centering
\resizebox{0.95\textwidth}{!}{
\begin{tabular}{c|c|ccccc|cccccc}
\toprule
\multirow{2}{*}{\textbf{Data Type}} & \multirow{2}{*}{\textbf{Dataset}} & \multicolumn{5}{c|}{\textbf{Handcrafted Text Prompts}}                                    & \multicolumn{6}{c}{\textbf{Learnable Text Prompts}}                          \\
                                    &                                       & \textbf{CLIP} & \textbf{CLIP-AC} & \textbf{WinCLIP} & \textbf{APRIL-GAN} & \textbf{AnoVL} & \textbf{CoOp} & \textbf{CoCoOp}  &  \textbf{BLIP}    & \textbf{AnomalyCLIP} & \textbf{FiLO} & \textbf{\coolname} \\ \midrule
\multirow{8}{*}{\textbf{Industrial}}          & \textbf{MVTecAD}                      & (38.4, 11.3)  & (38.2, 11.6)     & (85.1, 64.6)     & (87.6, 44.0)       & (89.8, 76.2)   & (33.3, 6.6)   & (86.7, 79.6)  & (85.1, 68.2) & ({\color[HTML]{0000FF}\textbf{91.1}}, {\color[HTML]{0000FF}\textbf{81.4}})  & ({\color[HTML]{FF0000}\textbf{92.3}}, 61.7)         & (90.6, {\color[HTML]{FF0000}\textbf{83.3}})  \\ 
% \hline
   & \textbf{VisA}                         & (46.6, 14.8)  & (47.8, 17.2)     & (79.6, 56.8)     & (94.2, 86.8)       & (89.9, 71.2)   & (24.1, 3.8)   & (93.6, 86.7)   & (85.7, 62.4)      & ({\color[HTML]{0000FF}\textbf{95.5}}, {\color[HTML]{0000FF}\textbf{87.0}})     & ({\color[HTML]{FF0000}\textbf{95.9}}, 85.4)     & ({\color[HTML]{FF0000}\textbf{95.9}}, {\color[HTML]{FF0000}\textbf{87.7}})   \\
                                    & \textbf{BTAD}                         & (30.6, 4.4)   & (32.8, 8.3)      & (72.7, 27.3)     & (60.8, 25.0)       & (93.2, 62.8)   & (28.6, 3.8)   & (86.1, 72.0)  & (89.3, 60.7)       & (94.2, {\color[HTML]{0000FF}\textbf{74.8}})  & ({\color[HTML]{0000FF}\textbf{94.5}}, 56.3)        & ({\color[HTML]{FF0000}\textbf{95.6}}, {\color[HTML]{FF0000}\textbf{75.1}})   \\ 
& \textbf{AITEX}                        & (53.2, 15.3)  & (47.3, 11.8)     & (62.5, 41.5)     & (78.2, {\color[HTML]{FF0000}\textbf{68.8}})       & (59.2, 49.1)   & (67.7, 54.9)  & (52.1, 56,9)  & (59.0, 57.7)       & ({\color[HTML]{FF0000}\textbf{83.0}}, {\color[HTML]{0000FF}\textbf{66.5}})    & (70.7, 25.3)      & ({\color[HTML]{0000FF}\textbf{82.0}}, 66.2)  \\
                                    & \textbf{DAGM}                         & (28.2, 2.9)   & (32.7, 4.8)      & (87.6, 65.7)     & (82.4, 66.2)       & (92.0, 78.8)   & (17.5, 2.1)   & (82.8, 75.1) & (90.0, 78.0)        & (95.6, {\color[HTML]{0000FF}\textbf{91.0}})   & ({\color[HTML]{0000FF}\textbf{96.8}}, 90.6)       & ({\color[HTML]{FF0000}\textbf{98.2}}, {\color[HTML]{FF0000}\textbf{95.0}})  \\
                                    & \textbf{DTD-Synthetic}                & (33.9, 12.5)  & (23.7, 5.5)      & (83.9, 57.8)     & (95.3, 86.9)       & (97.5, 90.4)   & (55.8, 36.0)  & (93.7, 83.7) & (96.3, 88.2)       & (97.9, {\color[HTML]{0000FF}\textbf{92.3}})   & ({\color[HTML]{0000FF}\textbf{98.1}}, 88.6)       & ({\color[HTML]{FF0000}\textbf{98.3}}, {\color[HTML]{FF0000}\textbf{93.3}})  
                                    
                                    \\
                                    & \textbf{SDD}                          & (28.4, 5.1)   & (33.5, 7.6)      & (95.9, 78.4)     & (93.0, 84.6)       & (97.9, 82.6)   & (91.8, 81.7)  & (93.7, 85.0)  & (95.6, 76.4)       & (98.1, {\color[HTML]{FF0000}\textbf{95.2}})   & ({\color[HTML]{FF0000}\textbf{99.0}}, 82.9)       & ({\color[HTML]{0000FF}\textbf{98.3}}, {\color[HTML]{0000FF}\textbf{94.1}})

                                    \\
                                    & \textbf{MPDD}                         & (62.1, 33.0)  & (58.7, 29.1)     & (76.4, 48.9)     & (94.1, 83.2)       & (84.0, 61.0)   & (15.4, 2.3)   & (95.2, 84.2) & (90.2, 69.4)       & ({\color[HTML]{FF0000}\textbf{96.5}}, {\color[HTML]{0000FF}\textbf{87.0}})  & ({\color[HTML]{0000FF}\textbf{95.7}}, 84.7)        & ({\color[HTML]{FF0000}\textbf{96.5}}, {\color[HTML]{FF0000}\textbf{87.9}})

                                    \\ \midrule
\multirow{6}{*}{\textbf{Medical}}   & \textbf{CVC-ColonDB}                  & (49.5, 15.8)  & (49.5, 11.5)     & (70.3, 32.5)     & (78.4, 64.6)       & (77.9, 49.8)   & (40.5, 2.6)   & (79.1, 69.7)  & (83.6, 66.9)     & ({\color[HTML]{0000FF}\textbf{81.9}}, {\color[HTML]{0000FF}\textbf{71.3}})    & (81.5, 63.9)      & ({\color[HTML]{FF0000}\textbf{85.0}}, {\color[HTML]{FF0000}\textbf{73.3}})  \\
                                    & \textbf{CVC-ClinicDB}                 & (47.5, 18.9)  & (48.5, 12.6)     & (51.2, 13.8)     & (80.5, 60.7)       & (82.1, 55.0)   & (34.8, 2.4)   & (83.4, {\color[HTML]{0000FF}\textbf{68.8}})        & (79.5, 63.1) &   (82.9, 67.8)    & ({\color[HTML]{0000FF}\textbf{84.6}}, 62.3)      & ({\color[HTML]{FF0000}\textbf{84.7}}, {\color[HTML]{FF0000}\textbf{70.1}})  \\
                                    & \textbf{Kvasir}                       & (44.6, 17.7)  & (45.0, 16.8)     & (69.7, 24.5)     & (75.0, 36.2)       & (72.5, 28.2)   & (44.1, 3.5)   & (79.1, 38.6)       & (76.5, 37.6)  & (78.9, 45.6)   & ({\color[HTML]{FF0000}\textbf{85.0}}, {\color[HTML]{FF0000}\textbf{53.2}})       & ({\color[HTML]{0000FF}\textbf{82.1}}, {\color[HTML]{0000FF}\textbf{49.9}})  \\
                                    & \textbf{Endo}                         & (45.2, 15.9)  & (46.6, 12.6)     & (68.2, 28.3)     & (81.9, 54.9)       & (80.5, 47.7)   & (40.6, 3.9)   & (83.1, 59.0)       & (78.5, 50.0)  & (84.1, 63.6)   & ({\color[HTML]{FF0000}\textbf{87.1}}, {\color[HTML]{0000FF}\textbf{65.6}})       & ({\color[HTML]{0000FF}\textbf{86.8}}, {\color[HTML]{FF0000}\textbf{67.6}})  \\
                                    & \textbf{ISIC}                         & (33.1, 5.8)   & (36.0, 7.7)      & (83.3, 55.1)     & (89.4, 77.2)       & (90.6, 79.8)   & (51.7, 15.9)  & (81.9, 68.9)      & (79.8, 57.6)   & ({\color[HTML]{0000FF}\textbf{89.7}}, 78.4)   & ({\color[HTML]{FF0000}\textbf{91.1}}, {\color[HTML]{0000FF}\textbf{80.1}})       & ({\color[HTML]{FF0000}\textbf{91.1}}, {\color[HTML]{FF0000}\textbf{81.6}})  \\
                                    & \textbf{TN3K}                         & (42.3, 7.3)   & (35.6, 5.2)      & (70.7, 39.8)     & (73.6, 37.8)       & (80.9, 50.5)   & (34.0, 9.5)   & (72.4, 41.0)       & (73.9, 44.3)  & ({\color[HTML]{0000FF}\textbf{81.5}}, {\color[HTML]{0000FF}\textbf{50.4}})   & (79.8, 48.5)       & ({\color[HTML]{FF0000}\textbf{84.7}}, {\color[HTML]{FF0000}\textbf{54.6}})  \\ \bottomrule
\end{tabular}
}
\caption{Pixel-level ZSAD results (AUROC, PRO) on 14 AD datasets. The best and second-best results are respectively highlighted in {\color[HTML]{FF0000}\textbf{red}} and {\color[HTML]{0000FF}\textbf{blue}}. Note that medical datasets in Table \ref{classification} do not have pixel-level ground truth. Thus, different medical datasets are used here. Detailed breakdown results for MVTecAD, VisA, DAGM, DTD-Synthetic, BTAD, and MPDD can be found in \texttt{Appendix} D.1.}
\label{segmentation}
\vspace{-0.3cm}
\end{table*}

\subsection{Main Results}

\noindent\textbf{Image-level ZSAD Performance.}
Table~\ref{classification} presents the image-level ZSAD results of \coolname, compared to ten SotA methods across 13 AD datasets, including nine industrial and four medical AD datasets. The results show that \coolname significantly outperforms the SotA models across almost all datasets. On average, compared to the best competing methods, it achieves up to 3.9\% AUROC and 3.0\% AP on industrial AD datasets and 2.3\% AUROC and 1.3\% AP on medical AD datasets. In particular, the weak performance of CLIP and CLIP-AC can be attributed to its over-simplified text prompt design. By utilizing more carefully designed handcrafted prompts, WinCLIP achieves better results than CLIP and CLIP-AC while preserving the training-free nature. 
APRIL-GAN and AnoVL improve over WinCLIP by using additional learnable layers and/or domain-aware tokens within the textual prompts. However, they heavily rely on sensitive handcrafted textual prompts and capture mainly coarse-grained semantics of abnormality, leading to poor performance on anomalies outside their seen domains (\eg, BTAD, MPDD, BrainMRI, HeadCT, Br35H).

As for learnable prompt methods, BLIP pretrained on a smaller scale of data, so it is less effective as in CLIP-related method. CoOp and CoCoOp are designed for general vision tasks, \ie, discriminating different objects, so they have weak capability in capturing the subtle differences between normality and abnormality on the same object. AnomalyCLIP and FiLO significantly improves ZSAD performance by learning textual prompts. 
However, their prompt design either fails to capture fine-grained abnormality details or relies heavily on external tools (LLMs and Grounding DINO). They also ignore crucial abnormality information that can be derived directly from the query input.
\coolname overcomes these limitations via the novel CAP and DAP.

\noindent\textbf{Pixel-level ZSAD Performance.}
We also compare the pixel-level ZSAD results of our \coolname with SotA methods across 14 AD datasets in Table~\ref{segmentation}. 
Similar observations can be derived as the image-level results. 
In particular, CLIP and CLIP-AC are the weakest among the handcrafted text prompt-based methods, primarily due to inappropriate text prompt designs. With better prompt engineering (and adaptation to AD in some cases), WinCLIP, APRIL-GAN, and AnoVL demonstrate better performance. 

For the learnable text prompt approaches, CoOp shows poor performance due to overfitting on the adaptation dataset, whereas CoCoOp mitigates this by introducing instance-conditional information, achieving substantial improvement over CoOp and competitive performance to BLIP, AnomalyCLIP and FiLO. \coolname exhibits superior performance in identifying a wide range of pixel-level anomalies, significantly outperforming SotA models across nearly all datasets, particularly in complex scenarios with larger sample sizes or requiring extensive knowledge transfer, such as DAGM and medical datasets. 
It surpasses the best competing methods by up to 1.4\% AUROC and 4.0\% AP on the industrial AD datasets, and by 3.2\% AUROC and 4.2\% AP on the medical AD datasets. This demonstrates the effectiveness of the fine-grained abnormality prompts in \coolname that adaptively capture detailed abnormality semantics in different datasets. 

\subsection{Analysis of \coolname}
\noindent\textbf{Module Ablation.}\label{module_ablation}
We conduct ablation study to assess the contribution of two modules of \coolname and report the averaged results across 18 industrial and medical datasets in Table~\ref{ablation}. AnomalyCLIP serves as our base model (\textbf{Baseline}), to which we separately add each of our proposed modules: `+ \textbf{CAP}' and `+ \textbf{DAP}'. 

As shown, applying CAP alone results in a significant improvement in image-level ZSAD performance due to its ability of learning the fine-grained abnormality details. To assess how important the orthogonal constraint loss ($\mathcal{L}_{oc}$) is in CAP, we further evaluate the performance with $\mathcal{L}_{oc}$ removed, denoting as `+ \textbf{CAP} w\textbackslash o $\mathcal{L}_{oc}$'. 
The results indicate that $\mathcal{L}_{oc}$ helps CAP work in a more effective way, justifying its effectiveness in encouraging the learning of complementary fine-grained abnormal patterns.

\begin{table}[t]
  \centering
  \resizebox{0.95\linewidth}{!}{
\begin{tabular}{c|cc|cc}
\toprule
\multirow{2}{*}{\textbf{Model}} & \multicolumn{2}{c|}{\textbf{Industrial Datasets}}       & \multicolumn{2}{c}{\textbf{Medical Datasets}}     \\
                       & \textbf{Image-level} & \textbf{Pixel-level} & \textbf{Image-level} & \textbf{Pixel-level} \\ \midrule
\textbf{Baseline}                 & (85.0, 83.6)         & (94.4, 84.8)     & (87.7, 90.6)         & (83.2, 62.9)    \\ \midrule
\textbf{+ CAP}          & ({\color[HTML]{0000FF}\textbf{88.1}}, {\color[HTML]{0000FF}\textbf{87.0}})         & (94.6, 83.9)    & ({\color[HTML]{0000FF}\textbf{90.6}}, {\color[HTML]{FF0000}\textbf{93.1}})         & (83.8, 63.8)             \\
\textbf{+ CAP w\textbackslash o $\mathcal{L}_{oc}$}            & (87.2, 86.3)         & (94.3, 83.5)     & (90.3, 91.8)         & (83.6, 63.8)  \\
\midrule

\textbf{+ DAP}       & (86.9, 85.2)         & ({\color[HTML]{0000FF}\textbf{94.8}}, {\color[HTML]{0000FF}\textbf{84.9}})      & (90.2, 92.3)         & ({\color[HTML]{0000FF}\textbf{84.6}}, {\color[HTML]{0000FF}\textbf{64.8}})            \\
\textbf{+ DAP w\textbackslash o $\mathcal{L}_{prior}$}               & (86.5, 85.1)         & (94.7, 83.7)    & (89.9, 92.3)         & (84.5, 64.3) \\
\midrule

\textbf{\coolname}        & ({\color[HTML]{FF0000}\textbf{88.5}}, {\color[HTML]{FF0000}\textbf{87.5}})         & ({\color[HTML]{FF0000}\textbf{95.0}}, {\color[HTML]{FF0000}\textbf{85.6}})         & ({\color[HTML]{FF0000}\textbf{91.0}}, {\color[HTML]{0000FF}\textbf{93.0}})         & ({\color[HTML]{FF0000}\textbf{85.7}}, {\color[HTML]{FF0000}\textbf{66.2}})     \\ \bottomrule
\end{tabular}
}
\caption{Image-level (AUROC, AP) and pixel-level (AUROC, PRO) results of ablation study. 
}
\label{ablation}
\vspace{-0.5cm}
\end{table}

When DAP is applied independently, it results in substantial improvements in both image- and pixel-level performance, with a more pronounced effect on medical datasets. This can be attributed to its ability of deriving data-dependent abnormality information from any target data to enhance the cross-dataset generalization of \coolname. Similarly, we assess the role of the abnormality prior selection loss ($\mathcal{L}_{prior}$) by removing it from DAP (`\textbf{DAP} w\textbackslash o $\mathcal{L}_{prior}$'). The results show that removing $\mathcal{L}_{prior}$ may introduce noisy priors from normal samples, leading to a significant performance drop. When all components are applied, the CAP and DAP complement each other in capturing high-quality abnormality semantics, resulting in the best ZSAD performance.

\noindent\textbf{Significance of Sample-wise Abnormality Prior in DAP.}
To evaluate the effectiveness of our sample-wise abnormality prior in DAP module, we further conduct an ablation study in which we modified the basis of the learned prior by substituting the selected abnormal patch token embeddings (\textbf{P}) with image token embeddings (\textbf{I}), as employed in CoCoOp. The results are presented in Table~\ref{image-prior}.
As shown, CoCoOP yields poor performance in both image- and pixel-level ZSAD tasks. This is primarily due to that the learned prior focuses more on class-level differences while ignoring the sample-specific abnormality details that are essential for effective AD. While the two variants of \coolname markedly improve performance over CoCoOP by filtering out normal sample-related noise and introducing fine-grained abnormality semantics through $\mathcal{L}_{prior}$ and CAP, they still fall short the performance of the full \coolname.

\vspace{-0.1cm}
\begin{table}[t]
  \centering
  \resizebox{0.95\linewidth}{!}{
\begin{tabular}{c|ccc|cccc}
\toprule
\multirow{2}{*}{\textbf{Model}} & \multirow{2}{*}{\textbf{Prior}} & \multirow{2}{*}{\textbf{$\mathcal{L}_{prior}$}} & \multirow{2}{*}{\textbf{CAP}}           & \multicolumn{2}{c|}{\textbf{Industrial}}                             & \multicolumn{2}{c}{\textbf{Medical}}                                \\
{} & {} & {} & {}      & \multicolumn{1}{c}{\textbf{image-level}}  & \multicolumn{1}{c|}{\textbf{pixel-level}}   & \multicolumn{1}{c}{\textbf{image-level}}  & \multicolumn{1}{l}{\textbf{pixel-level}}  \\ \midrule
\multicolumn{1}{l|}{\textbf{CoCoOP}} & \textbf{I}            & $\times$            & \multicolumn{1}{c|}{$\times$}  & (73.1, 70.1)                     & \multicolumn{1}{c|}{(85.5, 80.9)}         & (79.2, 83.5)                     & (79.8, 57.7)                     \\
\multicolumn{1}{l|}{\textbf{$\text{FAPrompt}^{--}$}} & \textbf{I}             & \checkmark            & \multicolumn{1}{c|}{$\times$}  & \multicolumn{1}{l}{(77.3, 75.2)} & \multicolumn{1}{c|}{(88.6, 81.0)}         & \multicolumn{1}{c}{(83.2, 84.8)} & \multicolumn{1}{c}{(81.6, 62.6)} \\
\multicolumn{1}{l|}{\textbf{$\text{FAPrompt}^{-}$}} &\textbf{I}             & \checkmark            & \multicolumn{1}{c|}{\checkmark}  & ({\color[HTML]{0000FF}\textbf{87.3}}, {\color[HTML]{0000FF}\textbf{86.4}})                     & \multicolumn{1}{c|}{({\color[HTML]{0000FF}\textbf{94.2}}, {\color[HTML]{0000FF}\textbf{83.5}})}         & ({\color[HTML]{0000FF}\textbf{88.4}}, {\color[HTML]{0000FF}\textbf{90.3}})                     & ({\color[HTML]{0000FF}\textbf{83.8}}, {\color[HTML]{0000FF}\textbf{65.5}}) \\

\multicolumn{1}{l|}{\textbf{\coolname}} & \textbf{P}             & \textbf{\checkmark}            & \textbf{\checkmark}      & ({\color[HTML]{FF0000}\textbf{88.5}}, {\color[HTML]{FF0000}\textbf{87.5}})         &  \multicolumn{1}{c|}{({\color[HTML]{FF0000}\textbf{95.0}}, {\color[HTML]{FF0000}\textbf{85.6}})}         & ({\color[HTML]{FF0000}\textbf{91.0}}, {\color[HTML]{FF0000}\textbf{93.0}})         & ({\color[HTML]{FF0000}\textbf{85.7}}, {\color[HTML]{FF0000}\textbf{66.2}})     \\ \bottomrule
\end{tabular}}
\caption{Image-level (AUROC, AP) and pixel-level (AUROC, PRO) results using different abnormality prior in DAP.
}
\label{image-prior}
\vspace{-0.5cm}
\end{table}

\noindent\textbf{\coolname vs Ensemble methods.}\label{ensemble_methods}
To learn more abnormalities, a straightforward solution is to ensemble existing ZSAD methods. Accordingly, we investigate two ensemble strategies in AnomalyCLIP for comparison: i) to learn an ensemble of AnomalyCLIP models with each model tuning an abnormality prompt on the auxiliary dataset using a different random seed (`AnomalyCLIP Ensemble'), and ii) to learn a AnomalyCLIP model with a set of multiple abnormality prompts constrained to be orthogonal via $\mathcal{L}_{oc}$ (`AnomalyCLIP Ensemble*'). As shown in Table~\ref{alternatives}, although both ensemble strategies yield performance gains over the base model AnomalyCLIP, the abnormality prompts derived from these methods are considerably less effective than those produced by \coolname. This shortfall is primarily because the ensemble approaches tend to generate highly redundant prompts, whereas \coolname learns complementary prompts that capture fine-grained abnormality details unattainable by the ensemble methods.

\begin{table}[t]
  \centering
  \resizebox{0.95\linewidth}{!}{
\begin{tabular}{c|cc|cc}
\toprule
\multirow{2}{*}{\textbf{Model}} & \multicolumn{2}{c|}{\textbf{Industrial Datasets}}       & \multicolumn{2}{c}{\textbf{Medical Datasets}}     \\
                       & \textbf{Image-level} & \textbf{Pixel-level} & \textbf{Image-level} & \textbf{Pixel-level} \\ \midrule
\textbf{AnomalyCLIP}                 & (85.0, 83.6)         & (94.4, 84.8)     & (87.7, 90.6)         & (83.2, 62.9)    \\ \midrule
\textbf{AnomalyCLIP Ensemble}          & (85.5, 84.0)         & ({\color[HTML]{0000FF}\textbf{94.7}}, {\color[HTML]{0000FF}\textbf{85.0}})  & (89.3, 91.3)         & (83.2, 62.4)   \\
        \textbf{AnomalyCLIP Ensemble*}           & (85.5, 82.6)         & (94.6, 84.5)  & (88.8, 91.0)         & (83.5, 65.6)   \\ \midrule

\textbf{Best-matched Prompts}          & (85.1, 84.9)         & ({\color[HTML]{FF0000}\textbf{95.0}}, 83.2)  & (85.6, 89.1)         & ({\color[HTML]{0000FF}\textbf{84.7}}, {\color[HTML]{0000FF}\textbf{65.9}})   \\
        \textbf{Weighted Prompts}           & ({\color[HTML]{0000FF}\textbf{87.7}}, {\color[HTML]{0000FF}\textbf{86.7}})         & (94.4, 83.2)  & ({\color[HTML]{0000FF}\textbf{90.9}}, {\color[HTML]{0000FF}\textbf{92.3}})         & (85.1, 65.7)   \\ \midrule
\textbf{\coolname}        & ({\color[HTML]{FF0000}\textbf{88.5}}, {\color[HTML]{FF0000}\textbf{87.5}})         & ({\color[HTML]{FF0000}\textbf{95.0}}, {\color[HTML]{FF0000}\textbf{85.6}})         & ({\color[HTML]{FF0000}\textbf{91.0}}, {\color[HTML]{FF0000}\textbf{93.0}})         & ({\color[HTML]{FF0000}\textbf{85.7}}, {\color[HTML]{FF0000}\textbf{66.2}})     \\ \bottomrule
\end{tabular}
}
\caption{Image-level (AUROC, AP) and pixel-level (AUROC, PRO) results comparison with alternatives.}
\label{alternatives}
\vspace{-0.5cm}
\end{table}

\noindent\textbf{Prototypical Abnormality Prompt vs Alternative Methods.} 
To verify the effectiveness of averaging the prompts to obtain a unified prototypical abnormality prompt, we compare two alternative variants of \coolname in Table~\ref{alternatives}: i) `Best-matched Prompts', which selects the most similar abnormality prompt per patch/image tokens during anomaly scoring, and ii) `Weighted Prompts', which assign importance weights to the abnormality prompts in the averaged operation based on the top $M$ abnormal patch tokens $\textbf{p}_x$.
The results indicate that `Best-matched Prompts' performs worse than simple averaging approach, as selecting only a single best-match prompt can lead to the loss of complementary information from other abnormality prompts, limiting the model's ability to detect the full spectrum of abnormalities. Although `Weighted Prompts' partially mitigates this issue by preserving complementary information through weighted aggregation, its performance still falls short of that achieved by simple averaging. This is due to that the weight learning network can overfit the auxiliary training dataset in the zero-shot setting. Thus, we adopt the straightforward averaging strategy in \coolname, as it effectively integrates diverse abnormality cues while preserving their complementary information.

\begin{figure}[t]
    \centering
    \includegraphics[width=\linewidth]{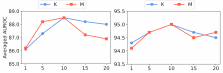}
    \caption{Image-level (\textbf{Left}) and Pixel-level (\textbf{Right}) AUROC results based on different value of $K$ and $M$.
    }
    \label{fig:para}
    \vspace{-0.5cm}
\end{figure}

\noindent\textbf{Hyperparameter Sensitivity Analysis.}
We analyze the sensitivity of two key hyperparameters of \coolname on industrial datasets in terms of image- and pixel-level AUROC results in Fig.~\ref{fig:para}, including the number of abnormality prompts $K$ in CAP and the number of selected patch tokens $M$ in DAP.
In particular, the performance gets improved with increasing $K$, typically peaking at $K = 10$. The performance may slightly declines when $K$ is chosen beyond 10. This suggests that while increasing the number of prompts helps capture a wider range of abnormalities, too large $K$ values may introduce noise or redundancy into the prompts.
As for the number of selected tokens, $M$, the performance exhibits a similar pattern, with the best performance obtained at $M = 10$. Beyond this point, a slight decline in performance is observed. This suggests that selecting too many abnormal patch candidates may introduce noise or less relevant patches into learned prior, leading to the learning of less effective fine-grained abnormalities (more comprehensive sensitivity analysis
can be found in \texttt{Appendix} C.4). 

\section{Conclusion}
In this paper, we propose a novel framework \coolname to enhance CLIP's performance in ZSAD by learning adaptive fine-grained abnormality semantics. \coolname introduces a Compound Abnormality Prompt learning (CAP) module that generates complementary abnormality prompts. It then incorporates a Data-dependent Abnormality Prior learning (DAP) module, which refines these prompts to improve cross-dataset generalization. The interaction between CAP and DAP enables the model to learn complementary fine-grained abnormality semantics. Extensive experiments on 19 datasets justify the effectiveness of \coolname.

\section*{Acknowledgments}

This research was supported in part by A*STAR under its MTC YIRG Grant (M24N8c0103), the Ministry of Education, Singapore under its Tier-1 Academic Research Fund (24-SIS-SMU-008), the Lee Kong Chian Fellowship (T050273), and the National Research Foundation (NRF), Singapore, through the AI Singapore Programme under the project titled “AI-based Urban Cooling Technology Development”(Award No. AISG3-TC-2024-014-SGKR).

{
    \small
    \bibliographystyle{ieeenat_fullname}
    \bibliography{main}
}

% \newpage
% \appendix
% \input{appendix}

\end{document}